\documentclass[conference]{IEEEtran}
\IEEEoverridecommandlockouts
\usepackage{cite}
\usepackage{amsmath,amssymb,amsfonts}
\usepackage{algorithmic}
\usepackage{graphicx}
\usepackage{textcomp}
\usepackage{xcolor}

\def\BibTeX{{\rm B\kern-.05em{\sc i\kern-.025em b}\kern-.08em
    T\kern-.1667em\lower.7ex\hbox{E}\kern-.125emX}}
\begin{document}

\title{Attention-Based Face AntiSpoofing of RGB Images, using a Minimal End-2-End Neural Network\\
}

\author{\IEEEauthorblockN{1\textsuperscript{st} Ali Ghofrani}
\IEEEauthorblockA{\textit{Lead AI Developer} \\
\textit{Alpha Reality Co}\\
Tehran, Iran \\
a.ghofrani@alphareality.io}
\and
\IEEEauthorblockN{2\textsuperscript{nd} Rahil Mahdian Toroghi}
\IEEEauthorblockA{\textit{Faculty of Media Technology and Engineering} \\
\textit{IRAN Broadcasting University (IRIBU)}\\
Tehran, Iran \\
mahdian.t.r@gmail.com}

\and
\IEEEauthorblockN{3\textsuperscript{rd} Seyed Mojtaba Tabatabaie}
\IEEEauthorblockA{\textit{CEO/CTO at Alpha Reality Co} \\
	\textit{AR/VR Solution Company}\\
	Tehran, Iran \\
	smtabatabaie@alphareality.io}

}

\maketitle

\begin{abstract}
Face anti-spoofing aims at identifying the real face, as well as the fake one, and gains a high attention in security sensitive applications, liveness detection, fingerprinting, and so on. 
In this paper, we address the anti-spoofing problem by proposing two end-to-end systems of convolutional neural networks. One model is developed based on the EfficientNet B0 network which has been modified in the final dense layers. The second one, is a very light model of the MobileNet V2, which has been contracted, modified and retrained efficiently on the data being created based on the Rose-Youtu dataset, for this purpose. The experiments show that, both of the proposed architectures achieve remarkable results on detecting the real and fake images of the face input data. The experiments clearly show that the heavy-weight model could be efficiently employed in server side implementations, whereas the low-weight model could be easily implemented on the hand-held devices and both perform perfectly well using merely RGB input images.

\end{abstract}


\section{Introduction}
Face anti-spoofing has always been a key challenging task of all face verification and recognition systems. Conventional face anti-spoofing systems used eigen faces~\cite{zhang1997face}, HoG (Histogram of Gradient)~\cite{albiol2008face}, or LBP (Local Binary Pattern) features to perform the task~\cite{rahim2013face}, whereas the recent systems mostly involve the deep neural features such as DeepFace~\cite{parkhi2015deep}, FaceNet~\cite{schroff2015facenet}, and OpenFace~\cite{baltruvsaitis2016openface}. 

Emerging and pervasive usage of the IR and depth sensors, has highly simplified the detection of real and fake images recently through some image processing techniques, however using of all these sensors is not always feasible and moreover the previously distributed products of most companies are not included with these technologies.

In this paper, we address the problem of face anti-spoofing through merely using the RGB frames of traditional cameras without using the auxiliary data, which is the most challenging task in this area.

In order to cope with the facial liveness detection challenge, several datasets with their specific properties have been released, so far. NUAA imposter~\cite{tan2010face}, which is publicly available, contains 7509 fake, as well as 5105 real images, however this volume of data, considering the general scale of the required data for deep learning purposes, is insufficient  and limiting. Casia-Surf, is another dataset recently published in CVPR 2019~\cite{Parkin_2019_CVPR_Workshops}, which contains all three types of data including RGB, IR, and depth data samples and is useful for multi-modal systems. This dataset contains 9608 training, as well as validation data samples.

The historically influential works in anti-spoofing area, contains four major approaches. One, is the texture-based methods which incorporate some hand-crafted features such as HoG, and LBP followed by traditional classifiers such as SVM to perform the task\cite{boulkenafet2016face,li2018face}. The temporal-based methods, on the other hand, either use the facial motion patterns (e.g., eye blinking) or involve the movements between face and the background and employ methods such as the optical flow to track the movement of face in order to discriminate real faces from the fake ones~\cite{inproceedings}. Some 3D structure-based methods have also been developed which either extract depth information from 2D images, or they analyze the 3D shape information being recorded with 3D sensors and then compare the 3D model of the input sample with that of a genuine face~\cite{8901932}. This method, however requires specific 3D devices which are not easily available and should be costly. Finally, the rPPG (Remote
Photoplethysmography) methods extract pulse signal
from facial videos without contacting any skin~\cite{lin2019face,song2018face}. Nevertheless, all these systems are highly vulnerable against the fake face attacks and masks, and may not cope with these attacks without the auxiliary data assistance, such as depth information, IR~\cite{liu2018learning}. In recent years, the deep learning-based methods have been pervasively used for many detection and recognition tasks, as well as anti-spoofing, ~\cite{chen2019face,chen2019attention}.

In this paper, we address the anti-spoofing problem and liveness detection using an end-2-end system. The novelty of this work is multi-fold: 1) No hand-crafted features (e.g., HoG, and LBP) have been utilized, 2) The proposed system requires no auxiliary data (e.g., Depth, IR) and resorts merely on the input RGB images, 3) Although the deep neural network structure has been employed, the entire system is light, portable and able to be deployed on hand-held devices and mobile phones with normal commercial processors, 4) The proposed system can deal with all types of fake images, such as depth-wise masks and high resolution display replaying, 5) Since each of the aforementioned achievements raises some challenging issues, the proposed system is able to cope with these issues, effectively.

The outlines of this paper is as follows. The next section, explains the background theory underlying the proposed system. Then, the experimental results and the analytics about them are pointed out. The conclusion terminates the paper and is followed by the references being cited in this paper.
\section{Related Works}
The problem of face anti-spoofing have been nearly solved for flagship devices. For example, the Face ID service on iPhone X, have been enabled to create a 3D mesh graph of the face, using Dot-projector, flood illumination, and IR sensors along with a dedicated neural network hardware (Neural Engine). Other brands, also use almost similar mechanisms to cope with the anti-spoofing problem. 

On the other hand, there are a handful of mid-range mobile handsets and previously sold out devices which lack these sensors and processing units. Furthermore, there are many verification tasks which are performed using laptop webcams which are totally missing these mechanisms. These issues motivate the work on a model which can address the problem using the RGB images, only.

\begin{figure}
\centering
\includegraphics[width=1\linewidth]{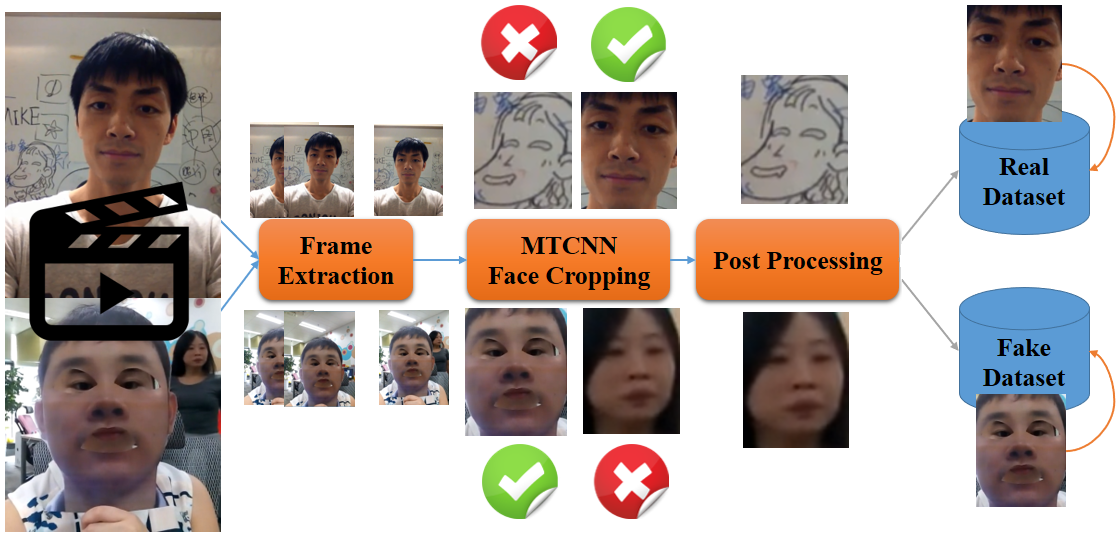}
\vspace{-6mm}
\caption{Data preparation flow graph, in order to gather the real and fake images from the dataset.}
\label{fig:2_big_picture}
\end{figure}

\begin{figure}[t]
	\centering
	\includegraphics[width=1\linewidth, height=.22\textheight]{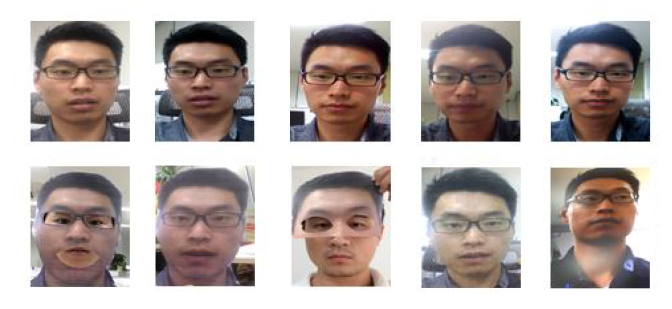}
	\vspace{-8mm}
	\caption{(Top) the real image samples, (Bottom) The fake image samples corresponding to the top images (from left to right): Mask with mouth and eyes cropped out, a paper mask without cropping, a paper mask with upper part cut in the middle, a paper print mask, and video replaying.}
	\label{fig:1}
\end{figure}

%\begin{figure}[!t]
%	\centering
%	\includegraphics[width=.85\linewidth, height=.25\textheight]{img/2_big_picture}
%	\caption{Data preparation flow graph, in order to gather the real and fake images from the dataset.}
%	\label{fig:2_big_picture}
%\end{figure}

\section{The Proposed Framework}

The problem of face anti-spoofing could be cast as a binary classification problem, which attempts to discriminate between real and fake images. However, the amount of fake samples is normally dominant vs. the real ones due to the enormous types of attacks and variations of the fake images within each type which could be given to the system. Hence, the system is likely to be exposed to the imbalanced training data. 

\subsection{Dataset Preparation Task}
In order to gather the required data for antispoofing purposes, there are issues which hinder the clean data preparation. For instance, the background person who passes by or the portraits in the background, could easily leak into the data if no preprocessing is performed. Correspondingly, these outliers have to be thrown away. The functional flow graph of the proposed system for appropriate and reliable data preparation is depicted in figure~\ref{fig:2_big_picture}.

Various datasets are generated to handle the liveness detection problem. NUAA imposter~\cite{tan2010face} which has been publicly available, contains only 7509 fake, and 5105 real images, and this volume of data for the deep network-based applications is not sufficient, at all. Casia Surf~\cite{zhang2019casia} is a recently released dataset, which contains the RGB, IR, and depth data samples, and is appropriate for multi-modal systems. It contains 9608 training, as well as validation data samples. This much data also do not suffice the deep structured training models. The dataset being used in this paper is ROSE-Youtu~\cite{li2018face}, which contains the real videos, and their corresponding fake videos. Therefore, the data samples are not extracted as images. Some samples of images taken from this dataset are depicted in figure~\ref{fig:1}.

 In order to extract the images from the ROSE-Youtu dataset, we employed the MTCNN network ~\cite{zhang2016joint} for face detection. To do so, we loaded every single video from the $3350$ videos in the dataset and spread it out into frames. Within each frame the face detection has been performed using MTCNN. Then, the face region has been cropped out, and put in its associated class of data~\cite{ghofrani2019realtime}.
 \begin{figure}[b]
 	\centering
 	\includegraphics[width=1\linewidth, height=.08\textheight]{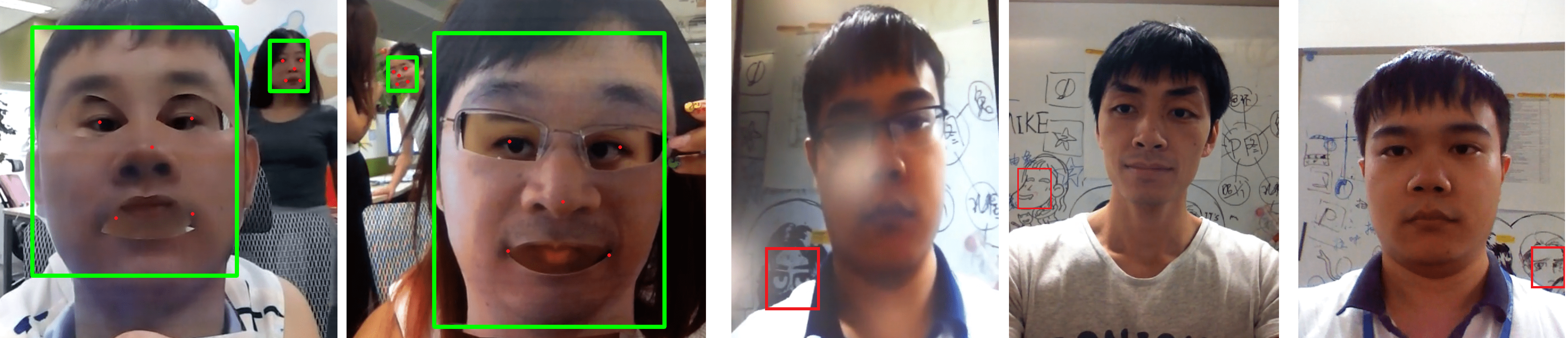}
 	\caption{Inappropriate samples (left to right): $1^{st}$ and $2^{nd}$ images: Fake samples with real background images being moved; Third up to fifth images: Fake samples with portrayed images in the background.}
 	\label{fig:3_bad_cases}
 \end{figure}
  Thus, we have prepared a set of $817,519$ data samples without augmentation being performed. During this preparation, we noticed some real images in which there are background people passing by, and this makes the real video to be classified as a fake one. In addition, there are videos which include portraits behind the scene, and this also causes the real video to be classified as fake. Samples of these cases are shown in figure~\ref{fig:3_bad_cases}.
 
In order to make sure that the model is robust against the person changing, the set of images related to a particular person has been extracted out of the data (which contains the samples from $20$ persons), and is used as the test data, for both the real and fake images. The rest of the dataset has been divided into $80\%$ and $20\%$ for training and validation purposes, respectively.
 
\subsection{The Proposed Architecture} 
\begin{figure}[t]
\centering
\includegraphics[width=.86\linewidth, height=.66\textheight]{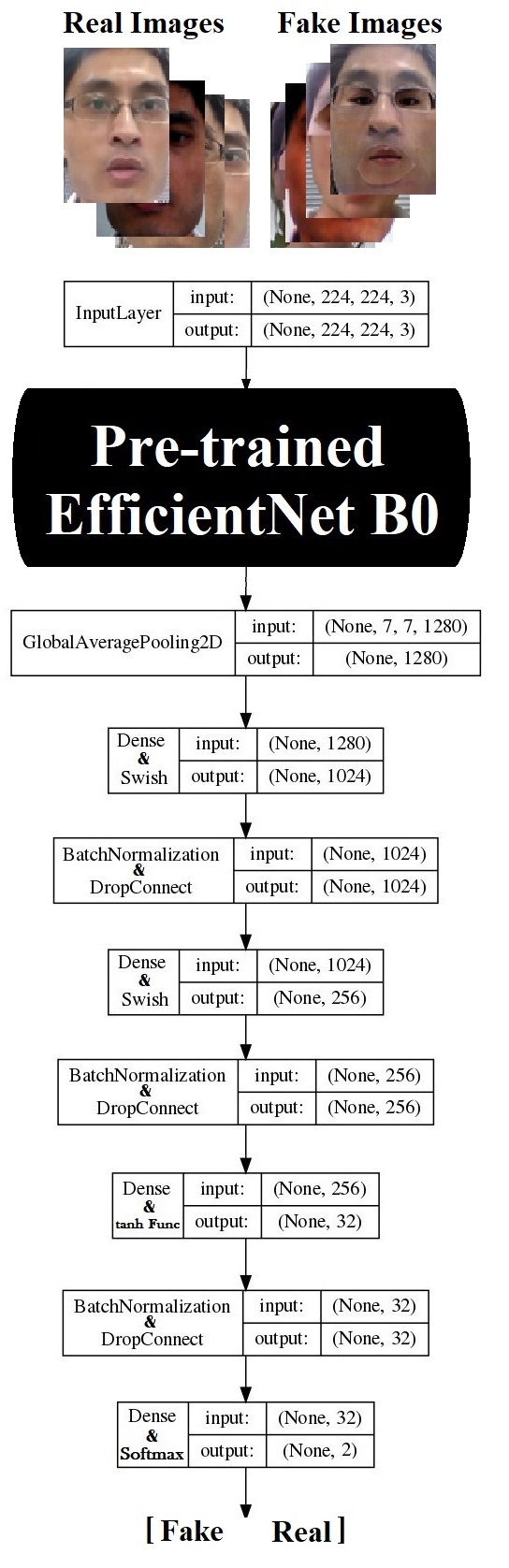}
\caption{Anti-spoofing architecture based on EfficientNet B0. Transfer learning has been involved for the weights of the network.}
\label{fig:4_eff}
\end{figure}

  After gathering and cleaning the data, in order to perform the binary classification (i.e., Real or fake images) a state-of-the-art EfficientNet architecture~\cite{tan2019efficientnet} has been employed, as depicted in figure~\ref{fig:4_eff}.
This network uses the B0 model, which has been pretrained on imagenet dataset, only as the initialization. The whole layers are trainable, and the stack of fully connected (FC) layers ($1024, 256, 32, 2$ neurons in each layer) with \textit{swish}~\cite{ramachandran2017swish} activation function in the first two FC layers, and the \textit{softmax} and \textit{tanh} activation functions for the final layers have been involved. Moreover, the \textit{dropconnect}~\cite{wan2013regularization} and \textit{batch normalization}~\cite{ioffe2015batch} has been performed between each two layers, to avoid overfitting. The entire number of parameters in the model is $5,592,606$, and the optimization method being employed is the \textit{Rectified-Adam}~\cite{liu2019variance}  minimization algorithm.
  
  Due to the fact that the ongoing binary classification is over the imbalanced data, as mentioned earlier, monitoring the accuracy of the network is not reasonable, the evaluations have been presented using the F1-score, which is a combination of precision and recall, as
  \begin{eqnarray}
  F_1\, \mathtt{score} = 2.\frac{precision.recall}{precision+recall}
  \end{eqnarray}
Furthermore, due to the classification nature of the problem the binary cross entropy has been chosen as the loss function.

During the training the following results, as in table~\ref{tb:t1}, have been achieved:
 \begin{table}[h]
 	\centering
 	\caption{Training and validation loss and F1-score for the architecture of figure~\ref{fig:4_eff}}
 	\begin{tabular}{|c|c|c|c|}
 		\hline 
 		\multicolumn{2}{|c|}{Train} & \multicolumn{2}{c|}{Validation}\tabularnewline
 		\hline 
 		\hline 
 		Loss & F1-score & Loss & F1-score\tabularnewline
 		\hline 
 		0.0204 & 99.37 & 0.0137 & 1.0\tabularnewline
 		\hline 
 	\end{tabular}
 	\label{tb:t1}
 \end{table}

\begin{figure}[t]
\centering
\includegraphics[width=1\linewidth, height=.13\textheight]{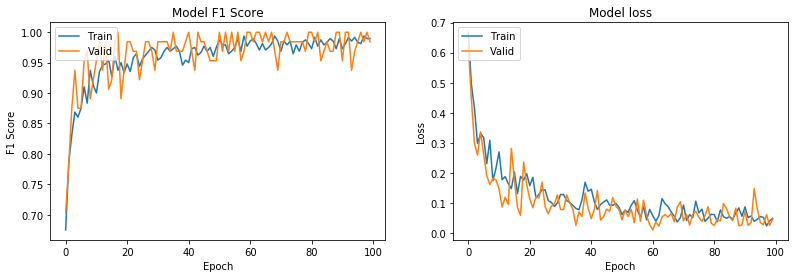}
\caption{Results of the F1-score, and loss of the training and validation for the architecture being proposed in figure~\ref{fig:4_eff}.}
\label{fig:5_eff_res}
\end{figure}

 Using dropconnect has caused the training loss to be less than the validation loss, as depicted in figure~\ref{fig:5_eff_res}.

\begin{figure}
\centering
\includegraphics[width=.85\linewidth, height=.2\textheight]{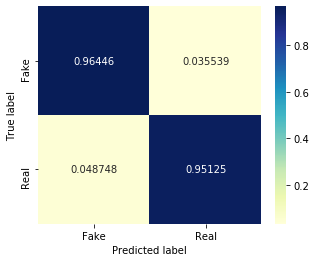}
\caption{The confusion matrix for the EfficientNet test data.}
\label{fig:6_eff_conf}
\vspace{-4mm}
\end{figure}

 The optimum parameters have been obtained after $97$ epochs, and the model needs $68,210$ MBytes to be saved. In addition, for the unseen test data the confusion matrix is, as in figure~\ref{fig:6_eff_conf}.
 
 As depicted in figure~\ref{fig:6_eff_conf}, the proposed model performs quite well, however due to the high number of parameters and using swish function it is not well suited for the client side implementations. This architecture, in stead is well qualified for the server side implementations.
 \subsection{Low Weight Model-Client Side}
\begin{figure}[b]
\centering
\includegraphics[width=1\linewidth, height=.2\textheight]{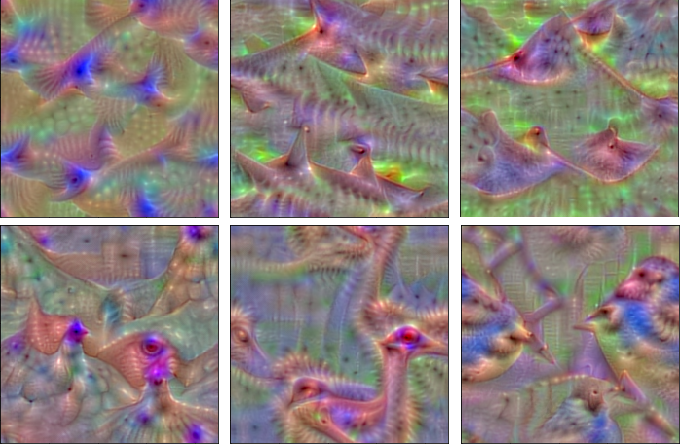}
\caption{Visualized final layer of the MobileNet. (Top-Left to right): Gold fish, White shark, Stingray, (Bottom-left to right): Hen, Oustrich, Brambling}
\label{fig:7_visual}
\end{figure}

  Another prevalent architecture which could be incorporated to perform the task is the MobileNet V2~\cite{sandler2018mobilenetv2} , which uses the separable CNN logic with depth-wise and point-wise (i..e Xception~\cite{chollet2017xception}). In this work, we have used a minimal structure of such a model, which uses the separable CNN logic with less number of parameters. A visual perception of the final layer of the MobileNet V2 being trained on the imageNet dataset has been depicted in figure~\ref{fig:7_visual}. In order to visualize the layer, the softmax has been omitted and the output has been activated using a linear activation.

\begin{figure}
	\centering
	\includegraphics[width=1\linewidth, height=.2\textheight]{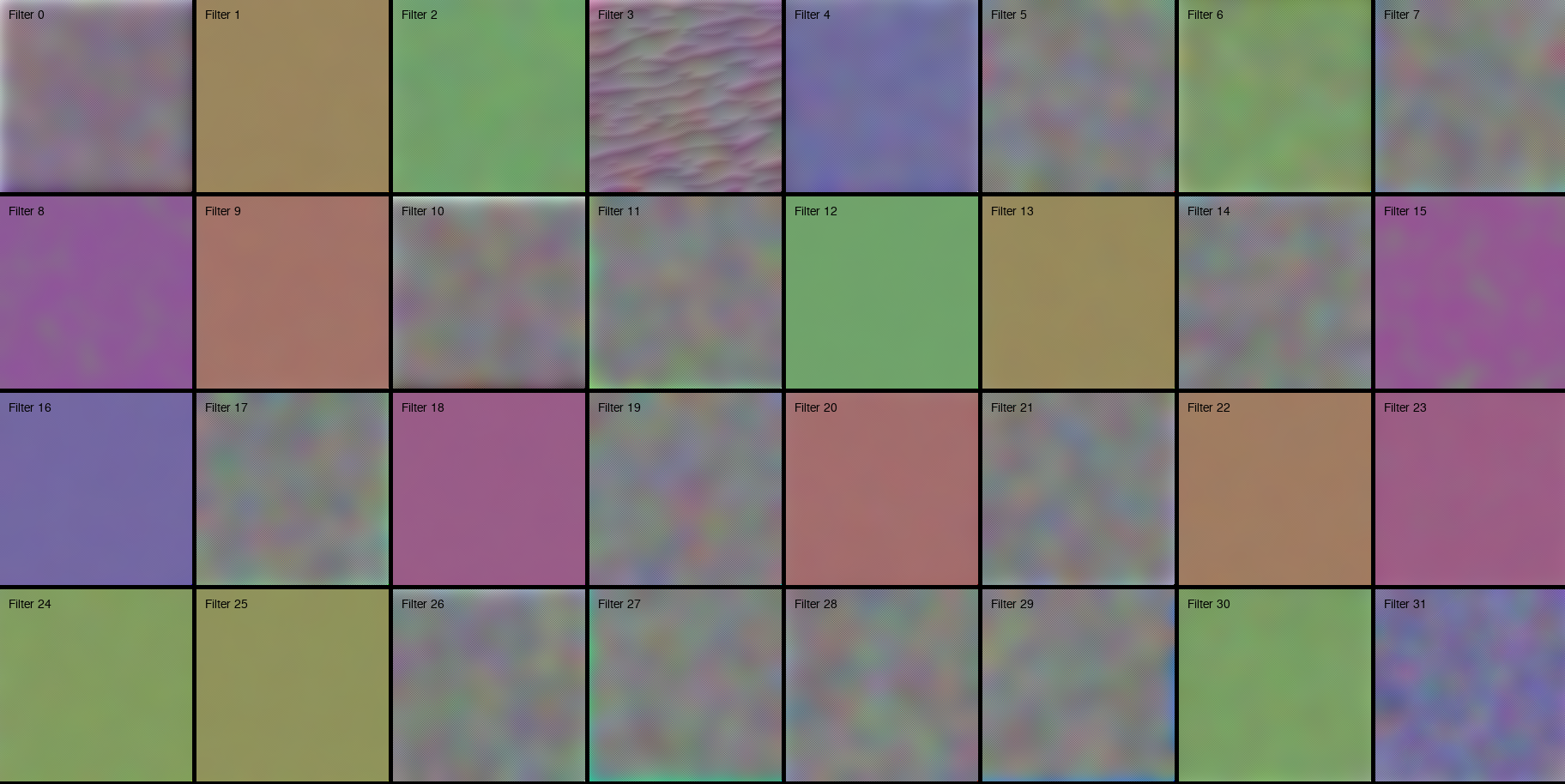}
	\vspace{-6mm}
	\caption{The primary level kernels for MobileNet V2, pretrained on imageNet.}
	\label{fig:8_first_cnn}
\end{figure}

\begin{figure}
\centering
\includegraphics[width=.85\linewidth, height=.64\textheight]{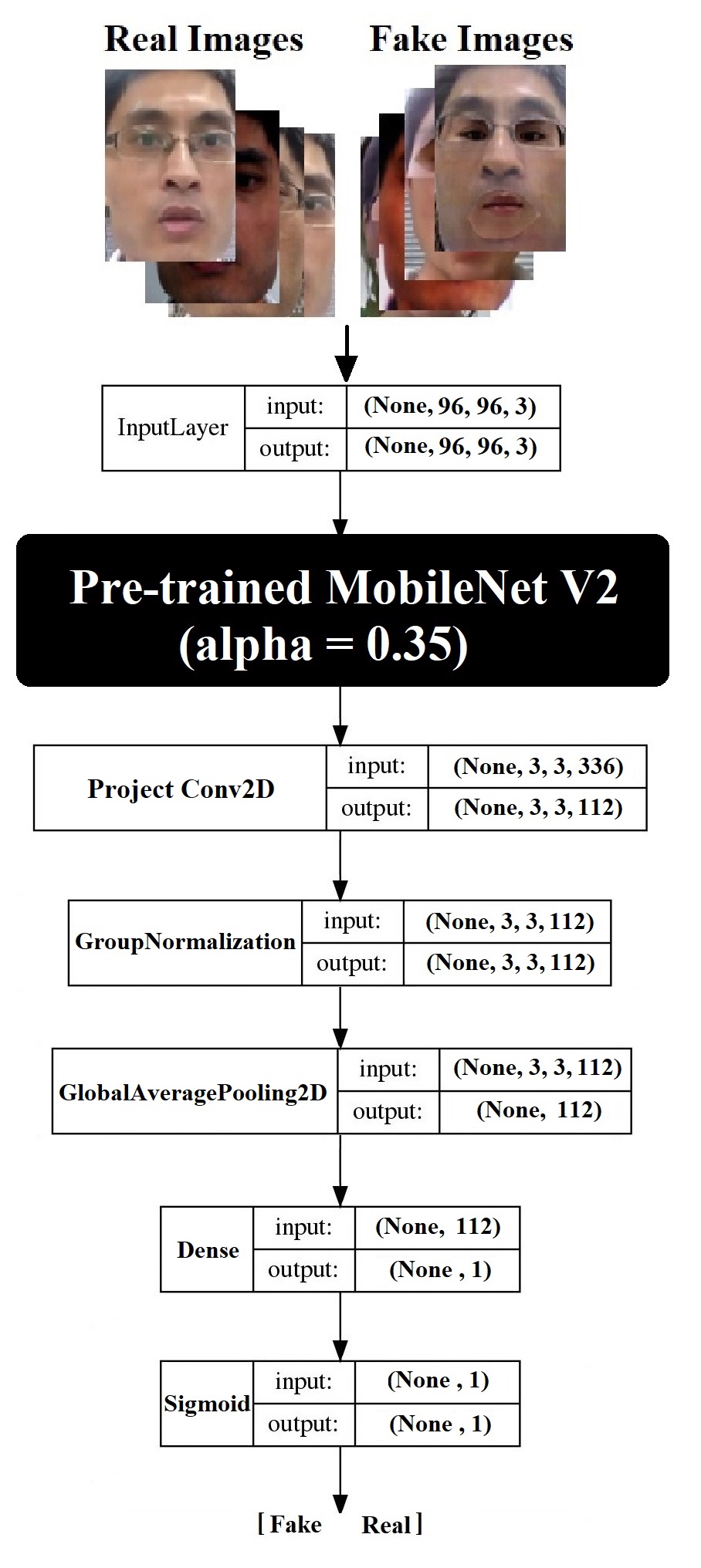}
\caption{Low-weight anti-spoofing architecture based on MobileNet V2. A contraction has been applied to the original network, and the remnant weights are used as the initializers.}
\label{fig:9_mobile}
\end{figure}

In addition to the previously shown output of the fully connected layer, we take a look at the primary convolution layer and some middle layers, as well. These figures are depicted in figure~\ref{fig:9_mobile}.

\begin{figure}
\centering
\includegraphics[width=1\linewidth, height=.24\textheight]{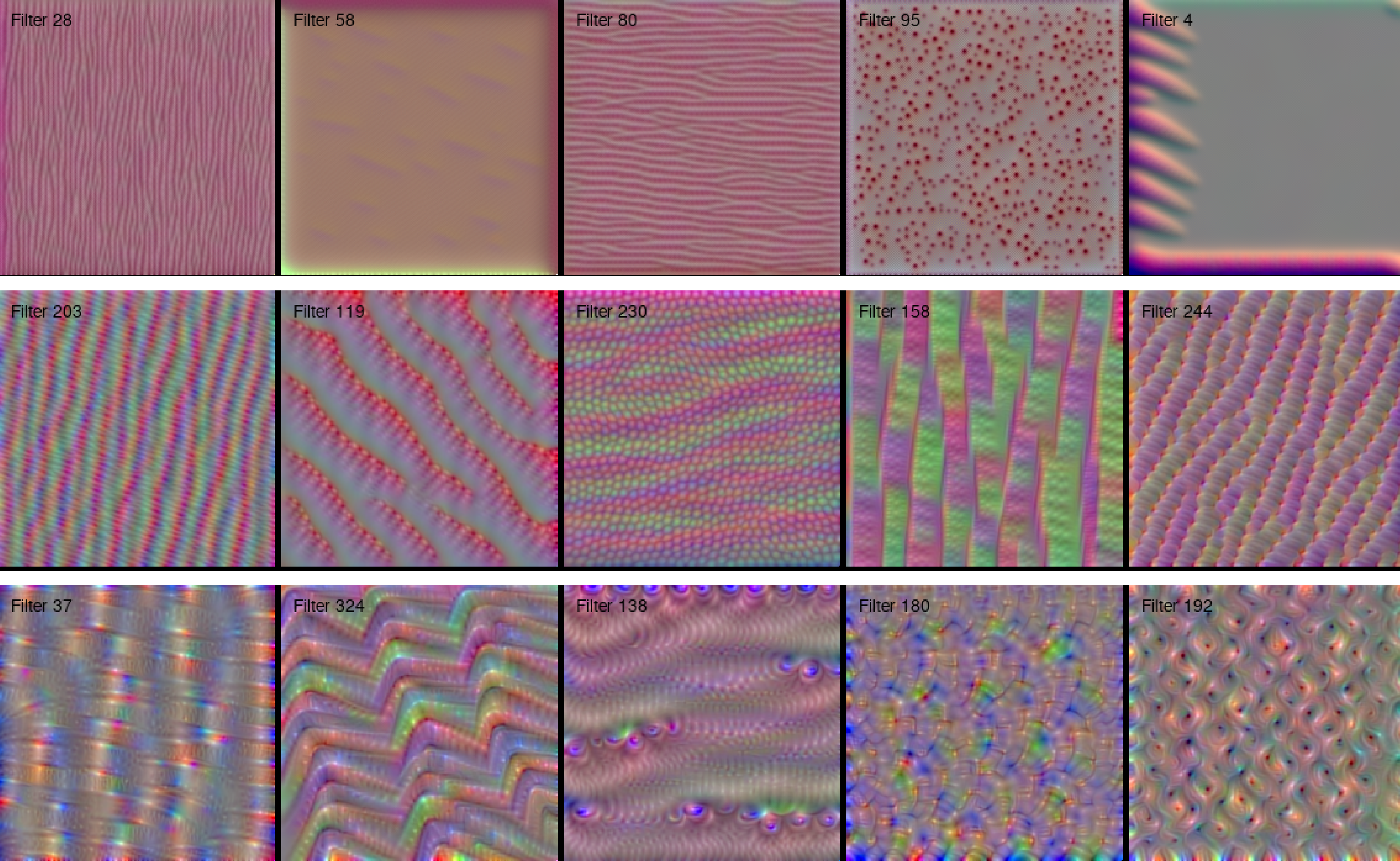}
\vspace{-6mm}
\caption{(Top to Bottom Rows): The low level kernels for MobileNet V2, The mid-level kernels for MobileNet V2, The high-level kernels for MobileNet V2, all of them are pretrained on imageNet dataset.}
\vspace{-1mm}
\label{fig:15_new}
\end{figure}

%
%\begin{figure}
%\centering
%\includegraphics[width=1\linewidth, height=.12\textheight]{img/imgss/middle1}
%\caption{The low level kernels for MobileNet V2, pretrained on imageNet.}
%\label{fig:middle1}
%\end{figure}
%\begin{figure}
%\centering
%\includegraphics[width=1\linewidth, height=.12\textheight]{img/imgss/middle2}
%\caption{The mid-level kernels for MobileNet V2, pretrained on imageNet.}
%\label{fig:middle2}
%\end{figure}
%\begin{figure}
%\centering
%\includegraphics[width=1\linewidth, height=.12\textheight]{"img/imgss/last_conv block"}
%\caption{The high-level kernels for MobileNet V2, pretrained on imageNet.}
%\label{fig:last_convblock}
%\end{figure}

Looking at these figures, indicates that the network has reached a high perception level for the imageNet dataset with numerous classes. However, in our work we are interested in only two classes, namely real and fake face images. Therefore, the basic network could be easily simplified with respect to the filters being used in convolution layers, and the parameters could be drastically decreased. In our proposed architecture, the number of the filters in each layer are $1/3$ of the original one, and the input size has been reduced to the minimum one, which is $96 \times 96 \times 3$. By applying these changes the model volume and the number of parameters has been changed from $16$ MBytes with $3.47$ million parameters into $3.7$ MBytes with $266,801$ parameters, respectively. Applying the deployment model conversion techniques, which has been proposed by Tensorflow (e.g., TF-lite conversion, and Quantization), even a more compact model volume could be achieved with $100$ KBytes.

\begin{figure}
\centering
\includegraphics[width=1\linewidth, height=.22\textheight]{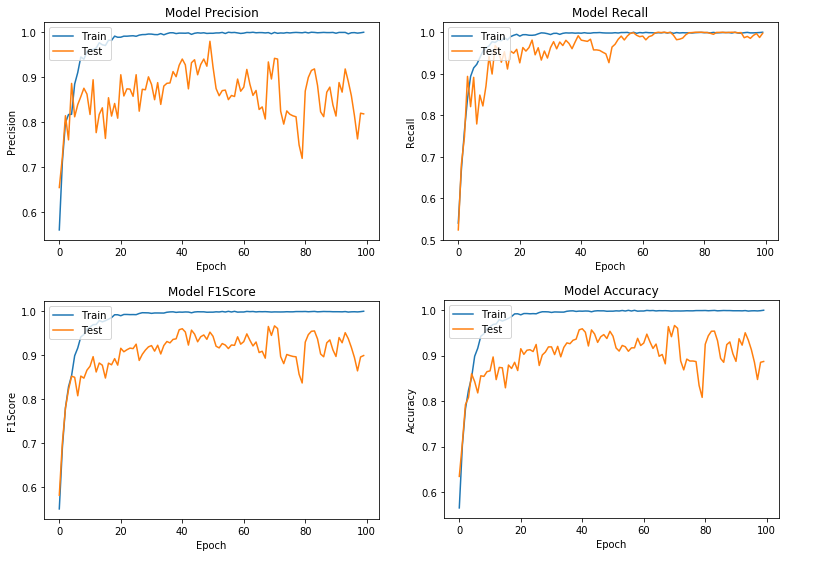}
\caption{The results of the low-weight proposed architecture of figure~\ref{fig:9_mobile} for training and validation data.}
\label{fig:10_mobile_res}
\end{figure}

In our implementation, we used these techniques, and using a GTX 1080 Graphic card, with $32$ GBytes of RAM, we could increase the batch size up to 718 samples, and for tuning the large batch the \textit{group normalizer}~\cite{wu2018group} has been employed. The evaluation results of training and testing of the proposed light model has been depicted in figure~\ref{fig:10_mobile_res}.

The achieved results are for $100$ epochs for each metric. 
  \begin{table}
  	\caption{Training and validation loss, accuracy, precision, recall, and F1 score of the proposed low-weight architecture.}
  \begin{tabular}{|c|c|c|c|c|c|c|c|c|c|}
  	\hline 
  	\multicolumn{5}{|c|}{Train} & \multicolumn{5}{c|}{Valid}\tabularnewline
  	\hline 
  	\hline 
  	loss & acc & pre & rec & F1 & loss & acc & pre & rec & F1\tabularnewline
  	\hline 
  	\scriptsize 0.0395 &\scriptsize 1.0 &\scriptsize 1.0 &\scriptsize 1.0 & \scriptsize 100 &\scriptsize 0.1654 &\scriptsize 96.29 &\scriptsize 97.81 &\scriptsize 100 &\scriptsize 95.69\tabularnewline
  	\hline 
  \end{tabular}
  \label{tb:t2}
  \end{table}
  
  \begin{figure}
\centering
\includegraphics[width=.84\linewidth, height=.18\textheight]{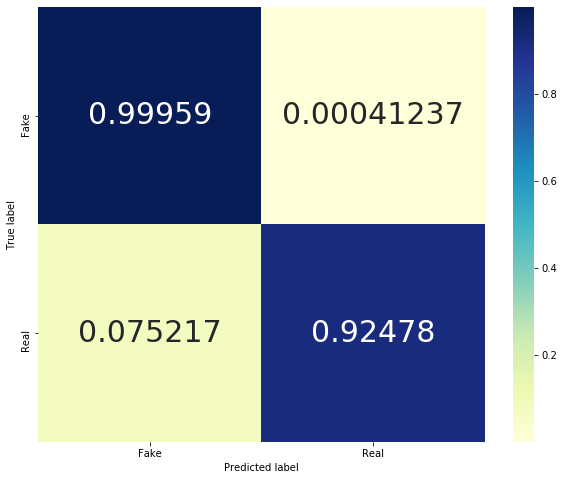}
\caption{The confusion matrix of the proposed low-weight proposed archiecture on the test data.}
\label{fig:11}
\end{figure}

By observing the visualized layers of the network, which has been trained for imageNet dataset containing $1000$ classes, and through induction we can logically dedicate that for our binary classification problem the low level kernels from the initial layers are not supposed to make a tremendous discriminative features, as opposed to the higher level layers, and therefore
they could be reduced. Thus, we canceled out half of the filters from the initial convolution layer, and a percentage of the rest. Figure~\ref{fig:8_first_cnn}, depicts the first layer kernels of MobileNet V2 network. Obviously, these number of kernels would not be very informative for a binary classification purpose. Thus, the network width controller coefficient of $0.35$ has been used in our experiments to achieve an optimum filter width within the MobileNet V2 network. Hence, the transfer learning is not exactly what we have performed in our work. We have contracted the pretrained MobileNet V2, then we used the initial weights of the contracted network, as depicted in figure~\ref{fig:15_new}, followed by our customized MLP stack ($336 \times 112 \times 1$ dense layers), with group normalization due to the huge batch size, as depicted in figure~\ref{fig:9_mobile}.
%
%
% as depicted in figures..... Hence, the most discriminative features are extracted from the final layer of the network, which is depicted in figure~\ref{fig:13}.
\begin{figure}
\centering
\includegraphics[width=1\linewidth, height=.1\textheight]{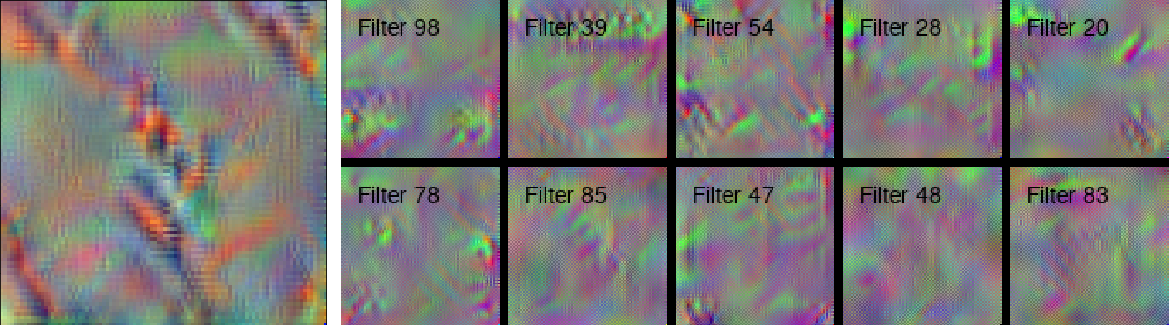}
\caption{(Left)The visualized dense layer of the proposed low-weight model, (The $2 \times 5$ matrix of images) kernels of the highest layer of the proposed architecture of figure~\ref{fig:9_mobile}}
\label{fig:13}
\end{figure}

%\begin{figure}
%\centering
%\includegraphics[width=1\linewidth, height=.2\textheight]{img/12}
%\caption{}
%\label{fig:12}
%\end{figure}
%\begin{figure}
%\centering
%\includegraphics[width=0.7\linewidth, height=.2\textheight]{img/13}
%\caption{}
%\label{fig:13}
%\end{figure}

\section{Experiments and Analytics}
 The confusion matrix of the low-weight anti-spoofing network over the ROSE-Youtu dataset is depicted in figure~\ref{fig:11}. The imbalanced nature of the data has impacted the real image detection outcome, compared to the EfficientNet B0 model, explained in the previous section. 
 
 As depicted in figure~\ref{fig:10_mobile_res}, the training process has been performed even faster than the original MobileNet model, since the number of parameters are dramatically decreased. However, the validation curve clearly shows a bias with respect to the training curve. The reason would be due to the tremendous reduction of the parameter numbers which pursue the network toward being underfitted. Identical to the figure~\ref{fig:7_visual}, in figure~\ref{fig:13} the final layer of our proposed network has been visualized, which works for the binary classification task, after the training phase is completed.
 
 The results of the proposed low-weight architecture, as depicted in figure~\ref{fig:10_mobile_res} and table~\ref{tb:t2}, obviously verifies its qualification for being used in the client side. 
 
\begin{figure}
\centering
\includegraphics[width=1\linewidth, height=.24\textheight]{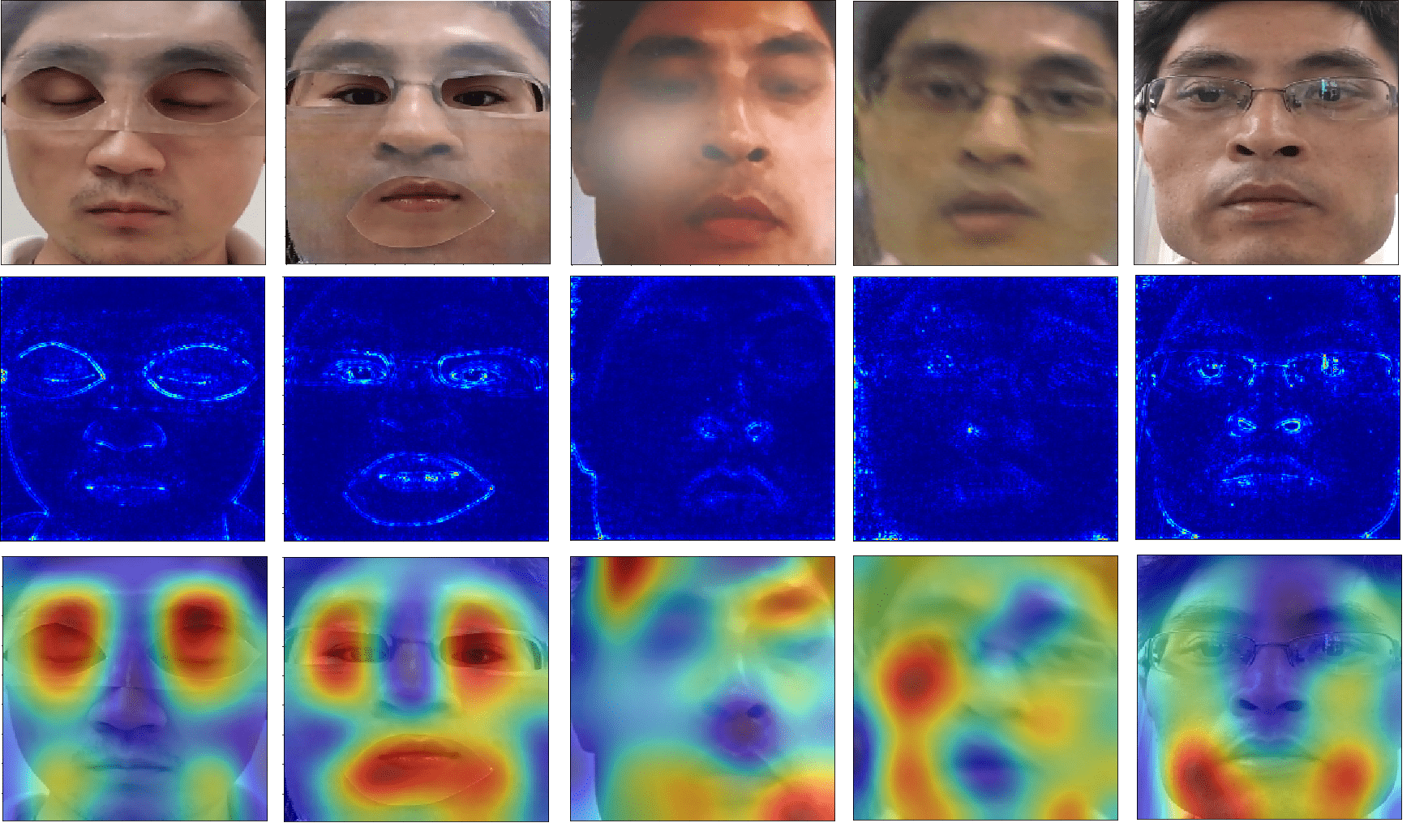}
\vspace{-6mm}
\caption{(Top-2-down rows): Test data (left to right: Upper mask, full mask, replay, Photo, real images); Saliency features of the test data~\cite{simonyan2013deep}; gradCAM attention visualization graph over the test data samples.}
\vspace{-4mm}
\label{fig:14}
\end{figure}

 As depicted in figure~\ref{fig:14}, the gradCAM attention visualization for the Up-mask image focuses on the eyes which has got an unusual depth, ~\cite{selvaraju2017grad}. For the full-mask image, both the eyes and mouth has grabbed the attention, and for the replay image and the photo image the attention distribution over the face has become scattered almost randomly. For the real face, however the attention is mostly on the chin, and distributed regularly.

 \vspace{-1mm}
\section{Conclusion} 
Two end-2-end attention-based face anti-spoofing models, have been proposed in this paper, one could be used for the server side and the other for the client side implementations, which merely incorporate the RGB images of the camera. These models require no auxiliary data (e.g., depth, IR) and perform remarkably well on the real and fake discrimination task.
The proposed model based on the EfficientNet B0, has performed perfectly well on the dataset, which enables it to be used in flagship mobile devices containing NPUs (dedicated Neural Processing Units), or in the server side.
 The proposed low-weight architecture requires very few number of parameters in a low volume which enables it to be efficiently used in mobile handsets. Various attacks have been experimented and both the heavy weight and the low-weight architectures have performed quite well on the fake data inputs which verify the robustness of the proposed models.

{\small
	\bibliographystyle{IEEEtran}
	\bibliography{MVIP2020_ghofrani}
}

\end{document}